\def\year{2020}\relax
\documentclass[letterpaper]{article} % DO NOT CHANGE THIS
\usepackage{aaai20}  % DO NOT CHANGE THIS
\usepackage{times}  % DO NOT CHANGE THIS
\usepackage{helvet} % DO NOT CHANGE THIS
\usepackage{courier}  % DO NOT CHANGE THIS
\usepackage[hyphens]{url}  % DO NOT CHANGE THIS
\usepackage{graphicx} % DO NOT CHANGE THIS
\usepackage{graphicx}  % DO NOT CHANGE THIS
\usepackage{amsmath}
\usepackage{amsfonts}
\usepackage{tabularx}

\title{Unpaired Image Enhancement\\
Featuring Reinforcement-Learning-Controlled Image Editing Software}
\author{Satoshi Kosugi and Toshihiko Yamasaki\\
Department of Information and Communication Engineering, The University of Tokyo, Tokyo, Japan\\
\{kosugi, yamasaki\}@hal.t.u-tokyo.ac.jp
}

\begin{document}

\maketitle

\begin{abstract}
  This paper tackles unpaired image enhancement,
  a task of learning a mapping function which transforms input images into enhanced images
  in the absence of input-output image pairs.
  Our method is based on generative adversarial networks (GANs),
  but instead of simply generating images with a neural network,
  we enhance images utilizing image editing software such as Adobe\textsuperscript{\textregistered} Photoshop\textsuperscript{\textregistered}
  for the following three benefits:
  enhanced images have no artifacts,
  the same enhancement can be applied to larger images,
  and the enhancement is interpretable.
  To incorporate image editing software into a GAN,
  we propose a reinforcement learning framework
  where the generator works as the agent that selects the software's parameters
  and is rewarded when it fools the discriminator.
  Our framework can use high-quality non-differentiable filters present in image editing software,
  which enables image enhancement with high performance.
  We apply the proposed method to two unpaired image enhancement tasks:
  photo enhancement and face beautification.
  Our experimental results demonstrate that the proposed method achieves better performance,
  compared to the performances of the state-of-the-art methods based on unpaired learning.
\end{abstract}

\section{Introduction}
Image enhancement is a task of learning a mapping function which transforms input images into enhanced images.
If we have a large number of original and enhanced image pairs, the task can be solved by image-to-image translation methods,
which have made significant progress~\cite{isola2017image,wang2018high} owing to the recent development of convolutional neural networks (CNNs).
However, in many cases, it is difficult to collect a large number of such image pairs.
To avoid this problem,
we address an image enhancement task that does not require paired image datasets; that is, unpaired image enhancement.
In this paper, we propose an unpaired image enhancement method which can be applied to real-world tasks.

A simple approach for unpaired image enhancement can be to use unpaired image-to-image translation methods,
which are mainly based on generative adversarial networks (GANs)~\cite{goodfellow2014generative}.
One of such methods is CycleGAN~\cite{zhu2017unpaired},
where generators which have an encoder-decoder architecture are trained with cycle consistency.
However, when using CNNs as decoders in real-world tasks,
there are three problems.
First, images generated by a CNN-based decoder have artifacts that can be attributed to CNN architecture.
Because artifacts can seriously degrade the quality of images, they can have fatal defects when used in practical applications.
Second, CNN-based decoders can only generate images with limited resolution in practice ({\it e.g.}, 512$\times$512px in the CycleGAN paper).
Recent high-resolution displays need 2000px or larger images,
but generating images with a high resolution makes the training unstable and time-consuming.
Third, image-to-image translation with CNN-based decoders is not interpretable.
Because the procedure is black-box, users cannot understand and manually adjust it.

To achieve unpaired image enhancement that is without artifacts, is scale-invariant, and is interpretable,
we use image editing software which edits the input image based on input parameters, such as Adobe Photoshop.
Using image editing software in the processing flow has the following three benefits:
edited images have no artifacts because the software is carefully designed for professional use,
the same editing can be applied to large sized images using the scale-invariant image editing filters provided by the software,
and the editing is interpretable allowing users to easily adjust it manually.
By using image editing software, we can achieve high-quality and highly practical image enhancement.
To utilize image editing software in a GAN,
we propose a reinforcement learning (RL) framework where the generator works as the agent controlling the software.
While a generator in a general GAN generates images directly,
the generator in our framework selects the software's parameters
and is rewarded when the edited result fools the discriminator.
By training the framework with RL, we can use high-quality non-differentiable image editing software.

To evaluate the performance of the proposed method, we apply it to two unpaired image enhancement tasks: photo enhancement and face beautification.
The experimental results show that the proposed method achieves better performance than previous approaches.

This paper makes the following contributions:
\begin{itemize}
 \item We achieve unpaired image enhancement that is without artifacts, is scale-invariant, and is also interpretable.
 \item We use image editing software and propose an RL framework to incorporate image editing software into a GAN.
 The generator is trained as the agent to select the software's parameters and is rewarded when it fools the discriminator.
 \item We apply the proposed framework to the tasks of photo enhancement and face beautification.
\end{itemize}

\begin{figure*}[t]
\centering
{\includegraphics[width=1\textwidth]{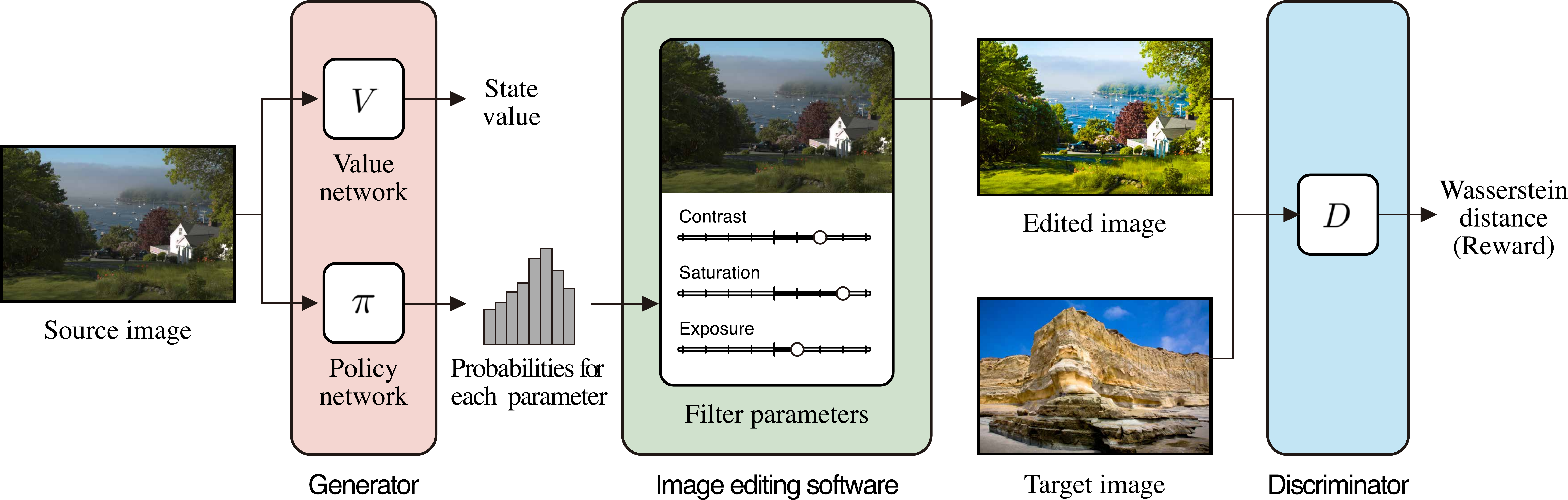}}
\caption{Overview of our method.
In our framework, the generator is trained with RL to control image editing software,
and the output of the discriminator is used as the reward.
\label{overview}
}
\end{figure*}

\section{Related Works}
\subsection{Image-to-Image Translation}
We formulate image enhancement as a task of learning the mapping from original images to images with the desired characteristics,
which is one of image-to-image translation problems.
A major CNN-based method for image-to-image translation is pix2pix~\cite{isola2017image},
which uses a conditional GAN~\cite{goodfellow2014generative} to learn a mapping from source to target images.
Based on this method, Wang et al.~\shortcite{wang2018high} achieved image-to-image translation with high resolution using multi-scale generators and discriminators.
These paired-learning methods require a large number of pairs of input and output images,
but in many cases, such pairs of images cannot be obtained.
To solve this problem, Zhu et al.~\shortcite{zhu2017unpaired} developed an unpaired image-to-image translation technique named CycleGAN,
where two GANs are trained using cycle consistency.
Kim et al.~\shortcite{kim2017learning} and Yi et al.~\shortcite{yi2017dualgan} also proposed similar methods and named them DiscoGAN and DualGAN, respectively.
Choi et al.~\shortcite{choi2018stargan} proposed a method named StarGAN that can handle translation between multiple domains.
We propose a more practical method than applying these methods directly to image enhancement.

\subsection{Reinforcement Learning for Image Processing}
In recent years, deep RL is being applied to image processing.
Cao et al.~\shortcite{cao2017attention} applied RL to super-resolution of facial images.
In that study, areas to be enhanced are sequentially selected by RL.
Li et al.~\shortcite{li2018a2} proposed an RL-based image cropping method,
where an agent sequentially updates the cropping window enabling high-speed cropping.
Yu et al.~\shortcite{yu2018crafting} used RL to select a toolchain from a toolbox for image restoration.
Furuta et al.~\shortcite{furuta2019fully} proposed a fully convolutional network that allows agents to perform pixel-wise manipulations.

One of the benefits of RL is that a framework containing non-differentiable functions
can be optimized.
Ganin et al.~\shortcite{ganin2018synthesizing} proposed a reinforced adversarial learning method
for synthesizing simple images of letters or digits using a non-differentiable renderer.
Because the image editing software we use and its renderer are both non-differentiable,
we apply some of their training strategy to our unpaired image enhancement method.

\subsection{Photo Enhancement}
Photo enhancement can be formulated as a translation between low-quality original images and high-quality expert-retouched images.
Bychkovsky et al.~\shortcite{bychkovsky2011learning} created a large-scale paired dataset for photo enhancement.
They hired five expert retouchers and created a collection of five sets of 5,000 input-output image pairs.
Using this paired dataset, Yan et al.~\shortcite{yan2016automatic} proposed an automatic photo adjustment framework,
which considers the local semantics of an image.
Gharbi et al.~\shortcite{gharbi2017deep} developed a CNN to predict the coefficients of a locally affine model in a bilateral space
and achieved high-speed edge-preserving photo enhancement.
Wang et al.~\shortcite{wang2019underexposed} built an underexposed image dataset and
proposed a network that can handle diverse lighting conditions.

Collecting pairs of original and expert-retouched images is labor-intensive.
To address this problem, unpaired learning methods
have been proposed.
Chen et al.~\shortcite{chen2018deep} made some improvements to CycleGAN to develop a stable two-way GAN framework.
Park et al.~\shortcite{park2018distort} created pseudo-input-retouched pairs by randomly distorting high-quality reference images.
Hu et al.~\shortcite{hu2018exposure} proposed a deep RL-based framework that applies retouching operations sequentially.
Their method is similar to our proposed method, but their architecture can only use differentiable filters.
While the available filters in their framework are limited,
our method can use a variety of filters because our method does not require filters to be differentiable.
In addition, the same framework can be applied to a completely different task such as face beautification.

\subsection{Face Attribute Manipulation}
Face beautification, a task of converting a less attractive face into an attractive face, is one application of face attribute manipulation.
On of the methods for face attribute manipulation is CycleGAN,
but the model is difficult to train, and generated images may include artifacts.
Several GAN-based approaches have been proposed to overcome this problem.
Shen et al.~\shortcite{shen2017learning} achieved efficient face attribute manipulation
by generating only the difference between images before and after the manipulation instead of generating the entire image.
Zhang et al.~\shortcite{zhang2018generative} introduced spatial attention to avoid edits in unrelated parts.

Another approach called deep feature interpolation (DFI), which does not use GANs, was proposed by Upchurch et al.~\shortcite{upchurch2017deep}.
By manipulating the deep features of the input image with a specific attribute vector and performing backpropagation to the image space,
the image after the manipulation can be obtained.
Using DFI, Chen et al.~\shortcite{chen2018facelet} achieved fast and high-quality face attribute manipulation with an end-to-end CNN that learns attribute vectors.
Chen et al.~\shortcite{chen2019semantic} developed a model that decomposes a facial attribute into multiple semantic components,
each corresponding to a specific face region.
These techniques have produced great results, but face attribute manipulation using CNNs inevitably generates artifacts.
This is a serious issue in face beautification. % , which is a task we deal with in this paper.

\section{Method}
Our goal is to learn a mapping function which transforms input images into enhanced images
in the absence of input-output image pairs.
We formulate this task as unpaired image-to-image translation from source domain $X$ to target domain $Y$,
where $X$ and $Y$ contain original images $\{{\bf x}_i\}^N_{i=1}$ and images with the desired characteristics $\{{\bf y}_j\}^M_{j=1}$, respectively.
We denote the data distribution as ${\bf x} \sim p_{s}$ and ${\bf y} \sim p_{t}$.
A simple approach can be training a CNN-based generator such as CycleGAN~\cite{zhu2017unpaired}.
However, CNN-based generators have several problems: the generated image has artifacts, the generator is not scale-invariant,
and the translation is not interpretable.
To achieve high-quality image enhancement
that addresses these problems, we introduce image editing software $\mathcal S$ such as Adobe Photoshop.
This image editing software $\mathcal S$ takes an image ${\bf x}$
and an action vector ${\bf a} = [a_1, a_2, ..., a_K]$ as input
and outputs the edited image ${\bf y^{\prime}} = {\mathcal S}({\bf x}, {\bf a})$.
Here, $K$ is the number of filters in the image editing software $\mathcal S$.
To incorporate the image editing software into a GAN,
we propose an RL framework,
which consists of the image editing software, one generator, and one discriminator.
In this framework, the generator works as an agent selecting parameters for the software
and is rewarded when it fools the discriminator.
Through the training process, the distribution defined by the generator ${\bf y^{\prime}} \sim p_{g}$ gradually approaches $p_{t}$.
We show the overview of our framework in Figure~\ref{overview} and give detailed explanations
of the discriminator and the generator in the following sections.

\subsection{Discriminator}
The training process of our discriminator $D$ is the same as that of discriminators in general GANs.
That is, it learns to distinguish the generated images from the real images.
We follow a method of Wasserstein GAN with gradient penalty (WGAN-GP)~\cite{gulrajani2017improved} and define the loss function as follows,
\begin{equation}
\mathcal{L}_{D}=-\mathbb{E}_{\mathbf{y} \sim p_{t}}[D(\mathbf{y})]+\mathbb{E}_{{\bf y^{\prime}} \sim p_{g}}[D(\mathbf{y}^{\prime})]+\lambda Z,
\end{equation}
where the first and the second terms increase the Wasserstein distance between generated images and real images.
$\lambda$ is a weight for $Z$, and $Z$ is a regularization term for the discriminator to stay in the set of Lipschitz continuous functions,
\begin{equation}
Z = {\mathbb{E}_{\hat{\mathbf{y}} \sim p_{\hat{\mathbf{y}}}}}[(\|\nabla_{\hat{\mathbf{y}}} D(\hat{\mathbf{y}})\|_{2}-1)^{2}].
\end{equation}
${\hat{\mathbf{y}}}$ is an image sampled along straight lines between images in $p_{t}$ and $p_{g}$.

\subsection{Generator}
We aim to incorporate image editing software into a GAN framework.
That is, our generator takes an original image ${\bf x}$ as input and outputs parameters for the software.
A simple approach is to design a differentiable image editing software $\overline{\mathcal S}$.
A generator $\overline{G}$ which generates parameters for $\overline{\mathcal S}$ can be directly optimized by minimizing the following loss:
\begin{equation}
\mathcal{L}_{\overline{G}} = - \mathbb{E}_{\mathbf{x} \sim p_{s}}[D(\overline{\mathcal S}(\mathbf{x}, \overline{G}(\mathbf{x})))].
\end{equation}
However, this method cannot use non-differentiable software such as Adobe Photoshop as $\overline{\mathcal S}$.

To utilize non-differentiable image editing software $\mathcal S$, we train the generator using RL.
In RL, an agent decides which action to execute according to the current state.
We define an original image ${\bf x}$ as the state and the parameter vector ${\bf a}$ as the action.
In the existing RL methods for image processing~\cite{cao2017attention,li2018a2,yu2018crafting,furuta2019fully,ganin2018synthesizing},
the agent receives operated images and decides actions sequentially, whereas
our agent receives an image and selects an action only once.
This is because $\mathcal S$ is not a linear function, for sequential actions ${\bf a}_1$ and ${\bf a}_2$,
\begin{equation}
\mathcal S(\mathbf{x}, {\bf a}_1 + {\bf a}_2) \neq　\mathcal S({\mathcal S(\mathbf{x}, {\bf a}_1)}, {\bf a}_2).
\end{equation}
Because it is hard for users to interpret sequential actions, we use only single-step actions.

We define the reward so that $\mathbf{y}^{\prime} = \mathcal S(\bf x, a)$
cannot be distinguished from images of the target domain $Y$.
The simplest reward is $D(\mathbf{y}^{\prime})$,
but maximizing only $D(\mathbf{y}^{\prime})$ can lead to lack of consistency between $\bf x$ and $\mathbf{y}^{\prime}$.
To deceive the discriminator with as small a change as possible, we define the reward $R$ as follows:
\begin{equation}
R = D(\mathbf{y}^{\prime}) - \alpha {\rm MSE}(\mathbf{x}, \mathbf{y}^{\prime}),
\end{equation}
where the second term calculates the mean squared error between two images.

We select advantage actor-critic (A2C)~\cite{mnih2016asynchronous} as a method of RL
following a training strategy by Ganin et al.~\shortcite{ganin2018synthesizing}.
A2C consists of value network $V$ and policy network $\pi$.
Value network $V({\bf x})$ is a module that estimates the value of the current state ${\bf x}$.
The loss to optimize $V$ is defined as follows:
\begin{equation}
{\mathcal L}_V = \left(V\left({\bf x}\right) - R\right)^2 / 2.
\end{equation}
Policy network $\pi(a_k|{\bf x})$ is a module that outputs the probability of each action $a_k$ in the current state $\bf x$ and is
trained to maximize the expected reward,
\begin{equation}
\mathcal{L}_{\pi}=\sum_{k}\left(-\log \pi\left(a_k| {\bf x}\right)\left(R-V\left({\bf x}\right)\right) - \beta H\left(\pi\left(a_k| {\bf x}\right)\right)\right).
\end{equation}
Intuitively, if the reward obtained by the operation ${\bf a} = [a_1, a_2, ..., a_K]$ is greater than the reward predicted by the value network,
the probability of ${\bf a}$ increases.
The second term is a function that calculates entropy,
which encourages the agent to explore and prevents convergence to local optima.

\subsection{Network Architecture}
\begin{figure}[t]
 \centering
{\includegraphics[width=1\columnwidth]{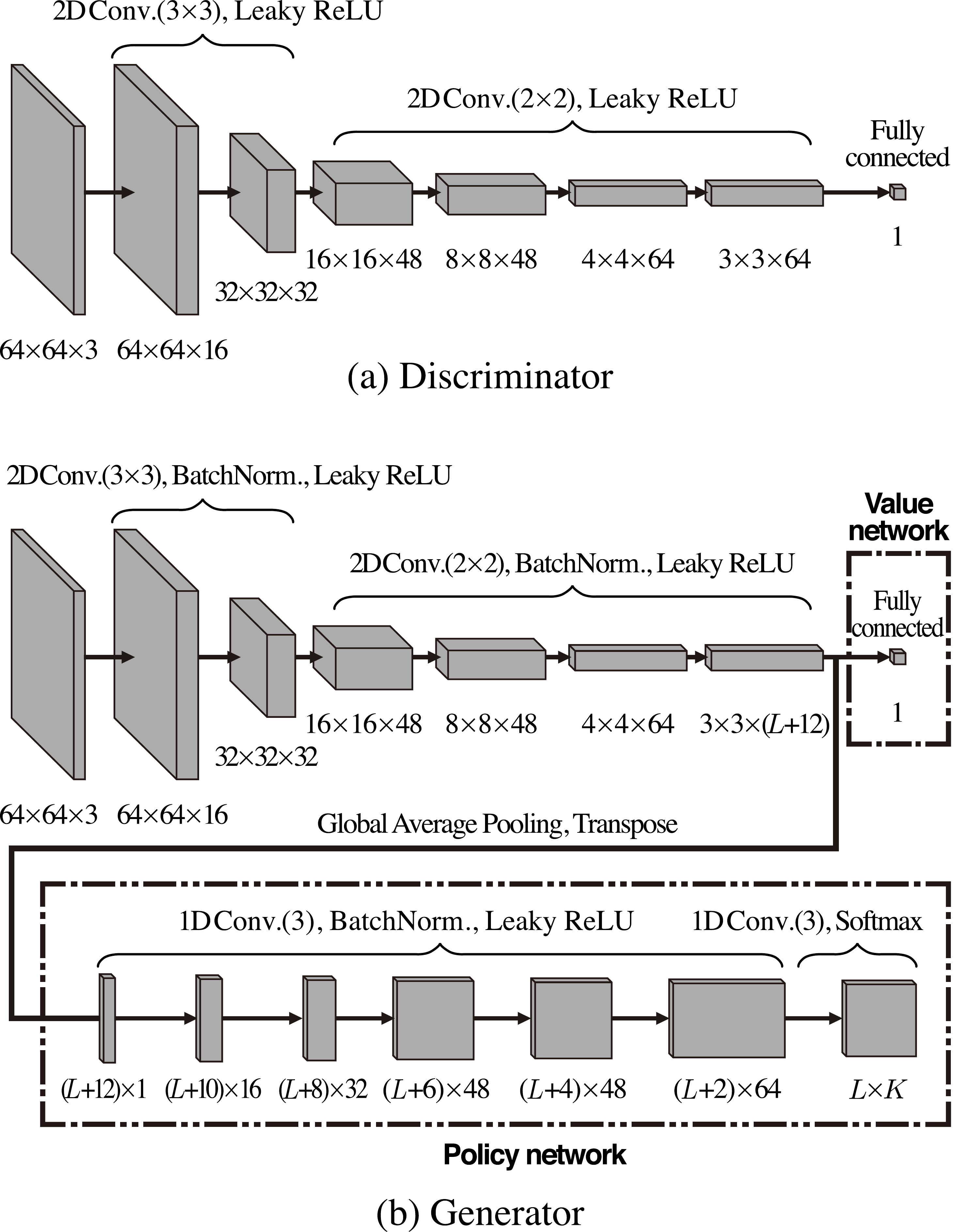}}
\caption{Network architecture of the discriminator and the generator.}
\label{network_architecture}
\end{figure}

In this paper, we use the discriminator and the generator whose architecture is shown in Figure~\ref{network_architecture}.
The discriminator has general CNN architecture similar to the one used in WGAN-GP~\cite{gulrajani2017improved}.
The generator consists of the policy network and the value network, which share the two-dimensional (2D) convolutional layers.

$\mathcal S$ can take continuous parameters, but an agent which selects continuous actions is hard to train.
Therefore, we design our agent to take discrete actions and the policy network to output probabilities for each discrete action.
We name the output of the policy network as ${\bf q}$, which is a matrix of $\mathbb{R}^{L\times K}$, and
$L$ is the number of the discrete steps of the parameters.
$\mathcal S$ has a maximum value $a^{max}_k$ and a minimum value $a^{min}_k$ for each $a_k$.
We divide the range between maximum and minimum values into $L$ steps, and
the policy network outputs probabilities for each discrete action as follows,
\begin{equation}
\pi\left(a^{min}_k + \left(a^{max}_k - a^{min}_k\right) \times \frac{l-1}{L-1}~\middle|~{\bf x}\right) = q_{lk},
\end{equation}
where $l\in\{1,2,...,L\}$.
To represent relationship between adjacent discrete steps ({\it e.g.}, $q_{lk}$ and $q_{(l+1)k}$),
we use one-dimensional (1D) convolutional layers to make the probabilities ${\bf q}$ from the CNN feature.
We do not use padding for the 1D convolutional layers,
because the padding can generate strange probability values at both ends of the steps
and can destabilize the training.

\begin{figure*}[t]
 \centering
{\includegraphics[width=1\textwidth]{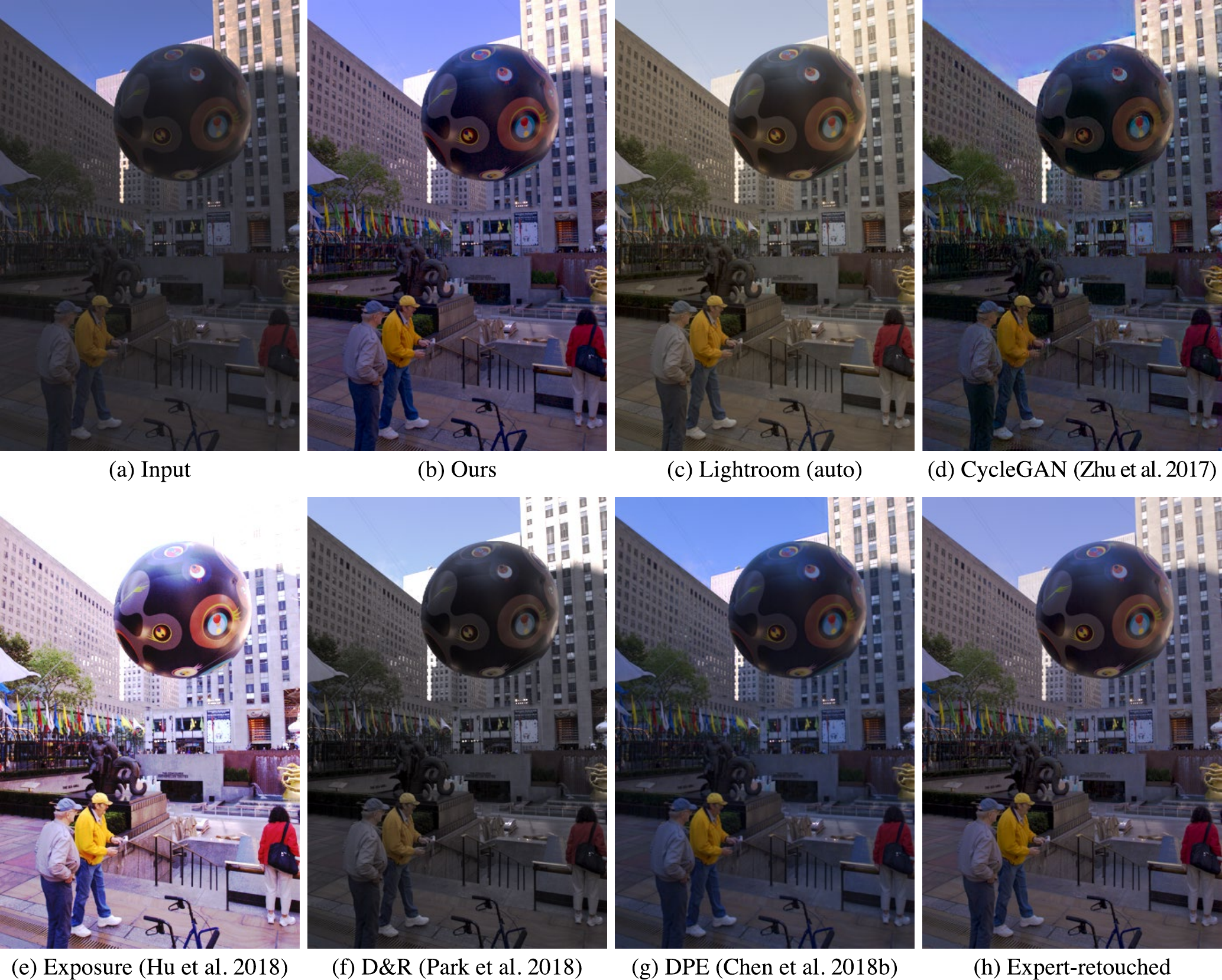}}
\caption{Qualitative comparison with other methods on a test image from the MIT-Adobe 5K dataset~\cite{bychkovsky2011learning}.}
\label{visual_result}
\end{figure*}

\begin{table}[t]
 \caption{The result of the quantitative comparison on the MIT-Adobe 5K dataset~\cite{bychkovsky2011learning}.}
{\includegraphics[width=1\columnwidth]{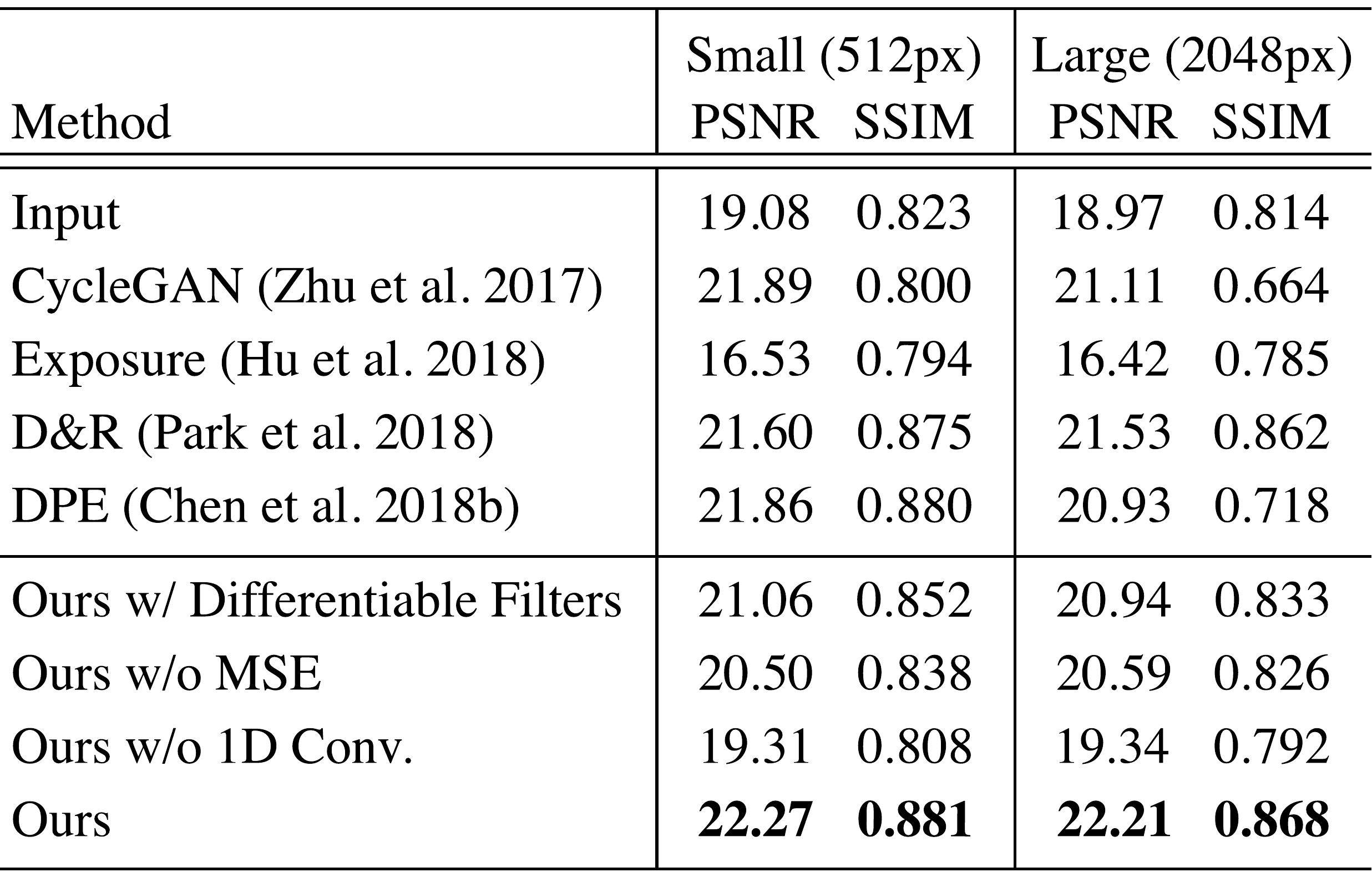}}
\label{psnr_ssim}
\end{table}

\begin{table}[t]
 \caption{The result of the user study on the MIT-Adobe 5K dataset~\cite{bychkovsky2011learning}.}
 \begin{center}
   {\includegraphics[width=0.7\columnwidth]{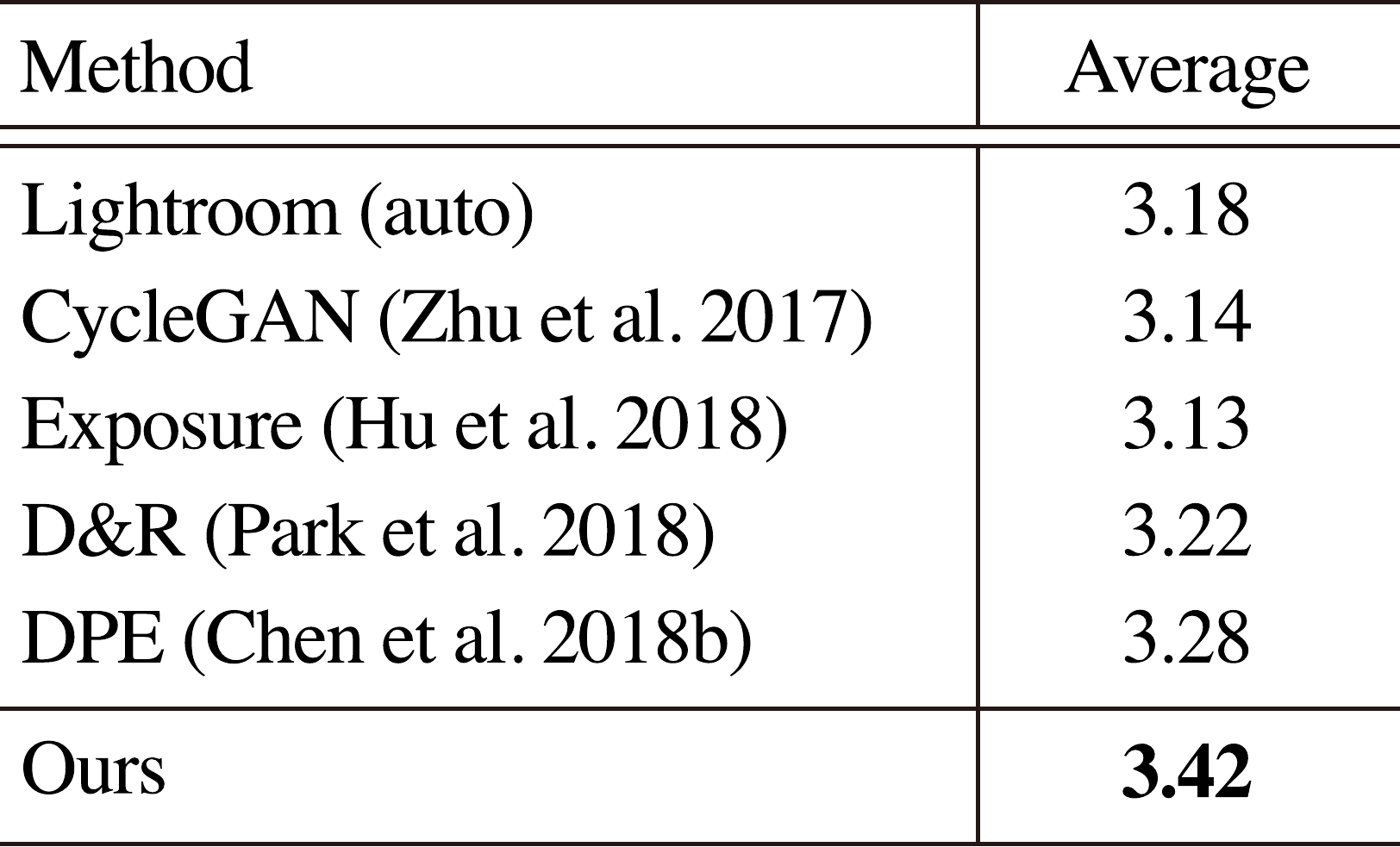}}
\end{center}
   \label{user_study}
\end{table}

\begin{figure}[t]
 \centering
{\includegraphics[width=1\columnwidth]{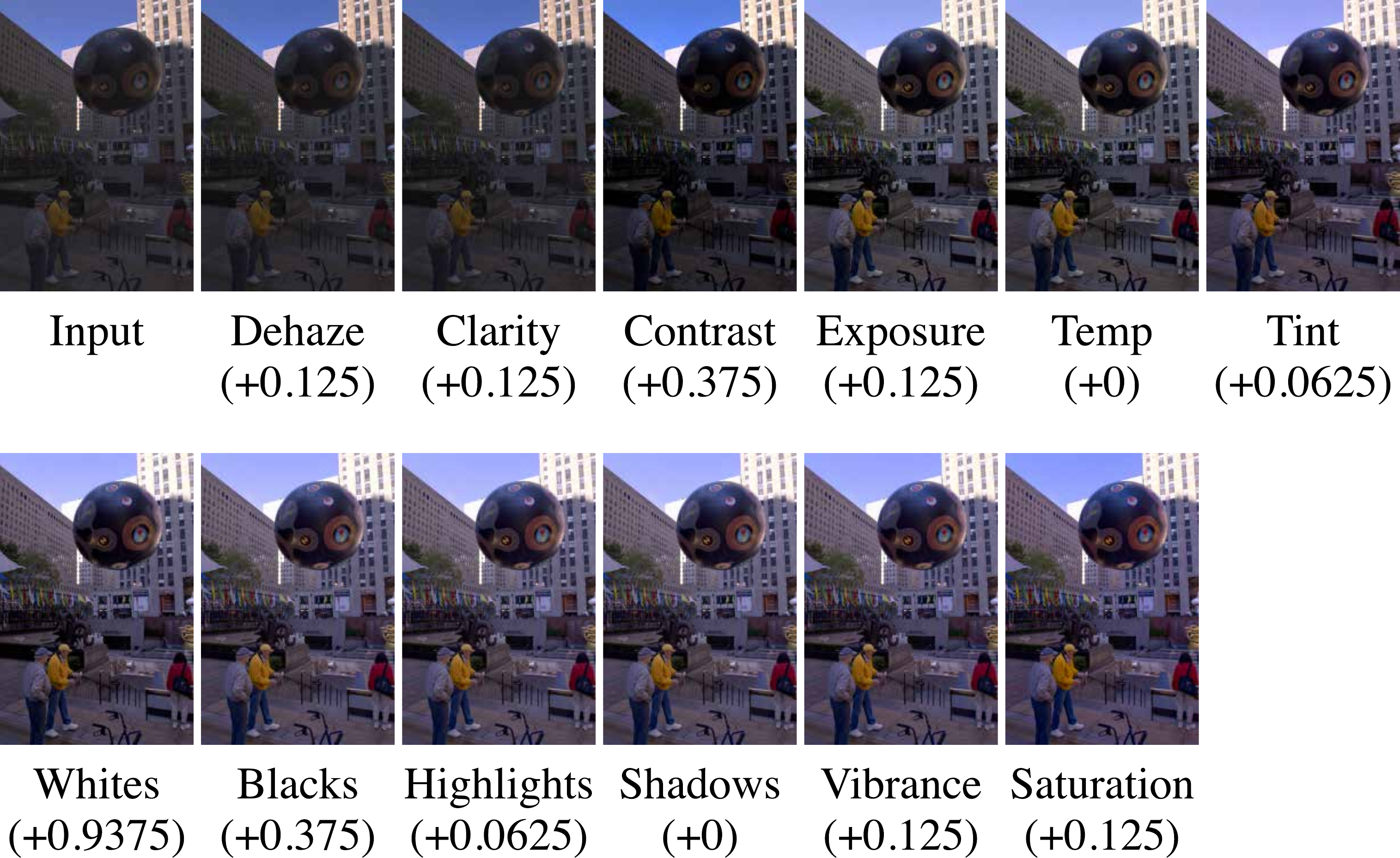}}
\caption{Application process of the filters. Values in parentheses are filter parameters, which are normalized to [-1, 1].\label{parameters}}
\end{figure}

\subsection{Train and Test}
While training, we resize all images to $64\times64$px,
and select the action ${\bf a}$ probabilistically according to $\pi$, that is,
\begin{equation}
a_{k} \sim \pi(a_k| {\bf x}).
\end{equation}
The resized image is edited according to ${\bf a}$.
While testing, the agent takes an image resized to $64\times64$px as a state and
selects action ${\bf a}$ deterministically,
\begin{equation}
l_{k} = {\rm arg}\max_{l}q_{lk},
\end{equation}
\begin{equation}
a_{k} = a^{min}_k + (a^{max}_k - a^{min}_k) \times \frac{l_k-1}{L-1}.
\end{equation}
Then, the selected action ${\bf a}$ is applied to the original image
because the operation of image editing software $\mathcal S$ is scale-invariant.

We train the discriminator and the generator alternately.
According to the paper of WGAN-GP~\cite{gulrajani2017improved},
the discriminator should be updated more frequently than the generator.
Following Ganin et al.'s~\shortcite{ganin2018synthesizing} training strategy,
we create a replay buffer which keeps images generated through the training process.
For every update of the generator, the discriminator is updated $U$ times using images
from the replay buffer.

\section{Experiments}
\subsection{Photo Enhancement}
\subsubsection{Dataset}
We apply the proposed method to photo enhancement,
a task of converting an original photo into an expert-retouched photo.
We use the MIT-Adobe 5K dataset~\cite{bychkovsky2011learning} for training and testing.
The dataset consists of 5,000 photos, and each image is retouched by five experts.
Following Chen et al.~\shortcite{chen2018deep}, we use the images retouched by Expert C as the target domain images.
To create unpaired image sets, we use 2,250 original images and non-overlapping 2,250 retouched images as training data,
and the other 500 pairs are used as test data.

\subsubsection{Implementation}
We choose Adobe Lightroom\textsuperscript{\textregistered} as the image editing software $\mathcal S$.
This tool can adjust the color, brightness, or contrast of an image by manipulating various filter parameters.
From the available filters, we choose the following: Dehaze, Clarity, Contrast, Exposure, Temp, Tint, Whites, Blacks,
Highlights, Shadows, Vibrance, and Saturation.
Because it is difficult to use Lightroom directly, we reproduce the filters on Python.
We optimize the discriminator and the generator using Adam~\cite{kingma2014adam} with
a learning rate of $10^{-4}$.
Other parameters $\lambda$, $\alpha$, $\beta$, $L$, and $U$ are 10, 100, 0.001, 33, and 5, respectively.

\subsubsection{Quantitative Evaluation}
We conduct a quantitative comparison with the existing methods.
We measure the difference between our result images and expert-retouched images using
two common metrics, {\it i.e.}, PSNR and SSIM.
In general, higher PSNR and SSIM mean better results.
To confirm that the proposed method is scale-invariant,
we conduct evaluations with small and large images
whose longer sides are 512px and 2048px, respectively.
We compare our method with CycleGAN~\cite{zhu2017unpaired} and some unpaired photo enhancement methods:
Exposure~\cite{hu2018exposure}, Distort-and-Recover (D\&R)~\cite{park2018distort},
and Deep Photo Enhancer (DPE)~\cite{chen2018deep}.
CycleGAN and DPE, which use CNN-based decoders, are trained using small images.
When testing with large sized images, small size results are resized to large size using
bicubic interpolation.
D\&R and Exposure, which are filter-based methods, can apply the same enhancement to
small and large images.

The result of the comparison is shown in Table~\ref{psnr_ssim}.
This result shows that our method achieves the best performance for both sizes.
DPE and the proposed method have almost the same values for SSIM with small size, with which the model is trained,
but DPE seriously drops SSIM on large sized images because the method is not scale-invariant.
D\&R and Exposure are filter-based methods and perform well for large images,
but the filters used in these methods are simple ones, resulting in scores lower than ours.
Compared to these filter-based methods, our method can use high-quality non-differentiable filters
and achieve image enhancement with high performance.

To analyze our method, we conduct ablation experiments.
First, we focus on the differentiability of the filters.
Our proposed framework is trained with RL,
which enables us to use non-differentiable filters in Lightroom.
To verify that the non-differentiable filters contribute to the high performance,
we replace them with differentiable filters used by Hu et al.~\shortcite{hu2018exposure} (Ours w/ Differentiable Filters).
As shown in the result, we obtain higher performance by using filters in Lightroom,
and the availability of non-differentiable filters is important to the high performance.

We also conduct experiments where we remove the mean squared error from the reward (Ours w/o MSE)
and replace the 1D convolutional layers with a fully connected layer (Ours w/o 1D Conv.).
The results show that the mean squared error and the 1D convolutional layers are
necessary factors for the high performance.

\subsubsection{Qualitative Evaluation}
We show a qualitative comparison with the other methods for a small sample in Figure~\ref{visual_result}.
In addition to the methods compared in the quantitative evaluation,
we use ``auto white-balance'' and ``auto-tone adjustment'' functions available in Adobe Lightroom,
which we name {\it Lightroom (auto)}.
As shown in this result, the Lightroom (auto) makes the color dull, CycleGAN generates
artifacts at the boundary between the sky and the building,
Exposure overexposes the image,
and D\&R outputs a slightly darker image than the target image.
Compared to these methods, our method can enhance the image without any artifacts
and properly reproduces the retouch by the expert.
DPE can achieve almost the same quality as ours
but is scale-sensitive as shown in the quantitative evaluation.

We show the sequential application process of the filters in Figure~\ref{parameters}.
Our proposed framework uses image editing software,
which enables users to interpret the enhancement and manually adjust it.
Note that although the filters are sequentially applied,
the agent selects all filter parameters at once.

\subsubsection{User Study}
We evaluate the proposed method through a user study.
We randomly select 100 original images from 500 test pairs and
perform enhancement using each existing method and the proposed method.
20 crowdworkers are hired via Amazon Mechanical Turk and
presented with 100 groups of results from existing and proposed methods,
which are arranged randomly to avoid bias.
Then, we ask the crowdworkers to give a five-grade rating from 1 (Bad) to 5 (Excellent).
Table~\ref{user_study} shows the average of all evaluations.
Our proposed method obtains higher evaluation than all the existing methods,
which shows that it is capable of a high-quality enhancement.

\begin{figure*}[t]
 \centering
{\includegraphics[width=1\textwidth]{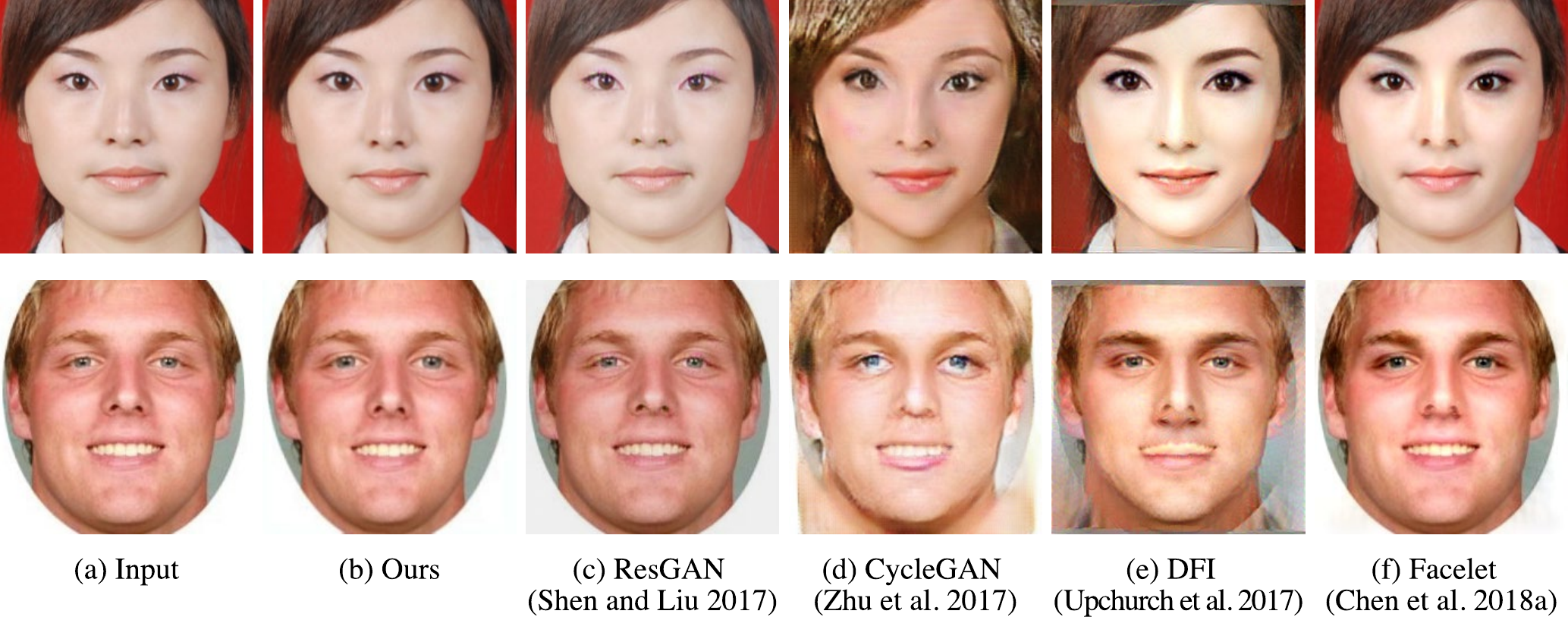}}
\caption{Qualitative comparisons with other methods on test images from the SCUT-FBP5500 dataset~\cite{liang2018scut}.}
\label{visual_result2}
\end{figure*}

\subsection{Face Beautification}
\subsubsection{Dataset}
We apply the proposed method to face beautification,
a task of converting a less attractive face into an attractive face.
For training and testing,
we use the SCUT-FBP5500 dataset~\cite{liang2018scut},
which has a total of 5,500 facial images and attractiveness scores within [1, 5].
We consider images with top 1,500 attractiveness scores as attractive images and the others as less attractive images.
Less attractive images with the lowest 1,500 attractiveness scores and all attractive images are used for the training,
and remaining less attractive images are used for the test.
We extract key points using the method of Kazemi et al.~\shortcite{kazemi2014one} to align face positions and resize images to 224$\times$224px.
The area outside of the face is masked-out with zero value while training to remove background information.

\subsubsection{Implementation}
For image editing software $\mathcal S$, we choose the Face-Aware Liquify function in Adobe Photoshop,
which provides filters to morph facial images by changing geometric structure such as eye size or face contour.
From the available filters, we choose the following: Eye Size, Nose Height, Nose Width, Upper Lip,
Lower Lip, Mouse Width, Mouse Height, Forehead, Chin Height, and Chin Contour.
Because it is difficult to use Adobe Photoshop directly, we reproduce the filters on Python.
The hyperparameters are the same as those used for photo enhancement, except that $\alpha$ and $L$ are 300 and 17, respectively.

\begin{table}[t]
 \caption{The result of the user study on the SCUT-FBP5500 dataset~\cite{liang2018scut}.}
 \begin{center}
   {\includegraphics[width=0.77\columnwidth]{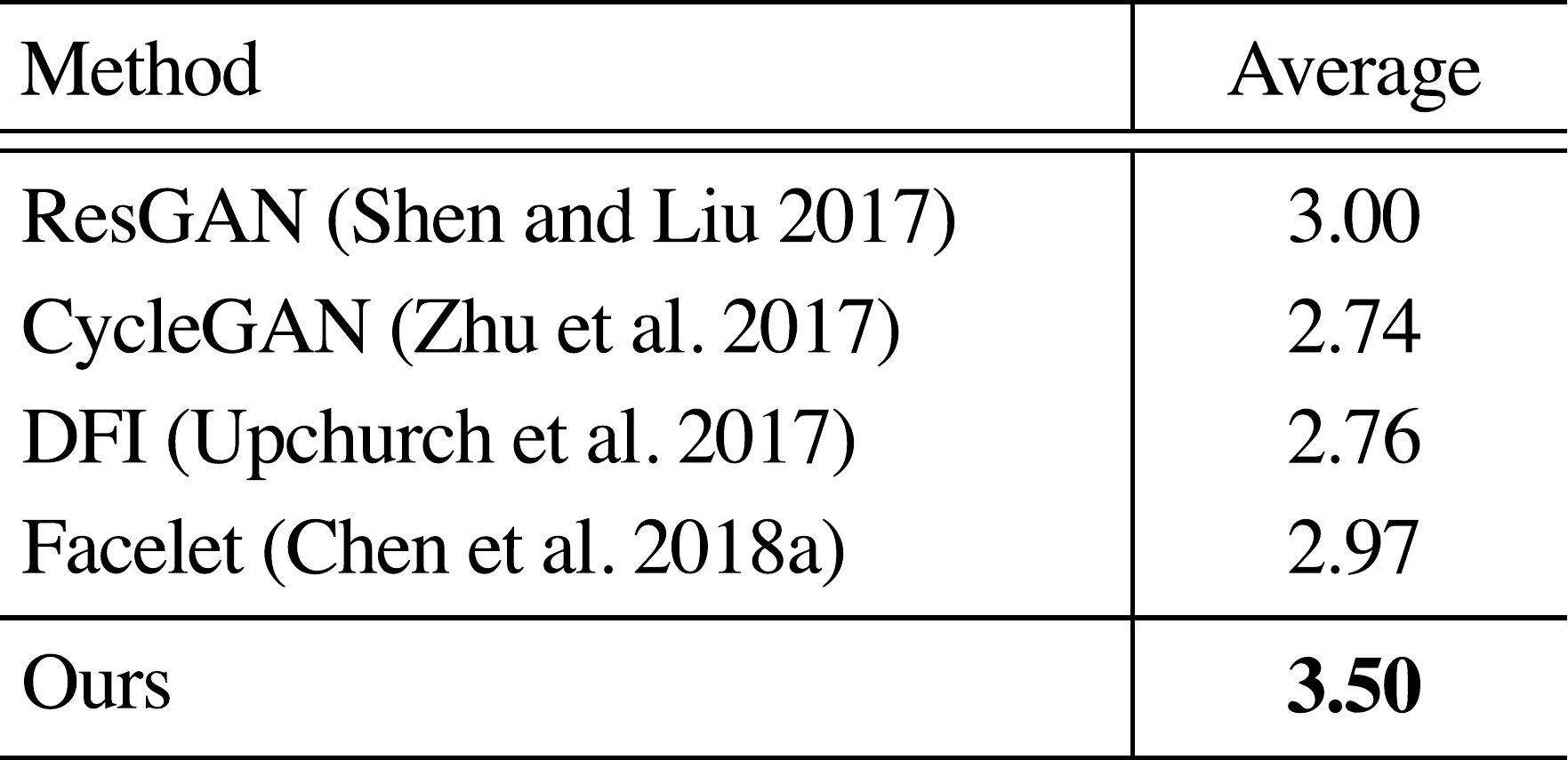}}
\end{center}
   \label{user_study2}
\end{table}

\subsubsection{Qualitative Evaluation}
In Figure~\ref{visual_result2}, we show qualitative comparisons with CycleGAN~\cite{zhu2017unpaired} and some face attribute manipulation methods:
ResGAN~\cite{shen2017learning}, DFI~\cite{upchurch2017deep}, and Facelet~\cite{chen2018facelet}.
All of these methods use CNN to manipulate portraits.
As shown in the results, ResGAN only generates artifacts around the eyes.
Although CycleGAN, DFI, and Facelet try to make the images look attractive,
the edited images have artifacts derived from the structure of CNNs,
which can prove fatal for the task of face beautification.
Compared to these methods, our method can naturally beautify the faces by
manipulating geometric structure such as enlarging the eyes or thinning the contours.

\subsubsection{User Study}
We evaluate the proposed method by a user study.
100 images are randomly selected
from less attractive images excluding those used for the training, and
we perform beautification using each existing method and the proposed method.
We ask crowdworkers to evaluate the images according to naturality and preference
in the same way as is done for photo enhancement.
Table~\ref{user_study2} shows the average of all evaluations.
The proposed method obtains higher evaluation than all existing methods,
which shows that our proposed method is capable of high-quality beautification.

\section{Conclusions}
In this study, we address unpaired image enhancement,
a task of learning a mapping function which transforms input images into enhanced images
in the absence of input-output image pairs.
Existing CNN-based methods have the following problems:
generated images have artifacts due to neural network architecture,
only images with limited resolution can be generated,
and the enhancement cannot be interpreted.
To solve these problems,
we use image editing software such as Adobe Photoshop,
which can perform high-quality enhancement and avoids the problems.
To use image editing software in a GAN, we propose an RL
framework where the generator works as an agent controlling the software
and the output of the discriminator is used as the reward.
The framework can use carefully designed non-differentiable filters,
which enable high-quality enhancement.
We apply the proposed method to photo enhancement and face beautification.
The experimental results show that our method performs better than existing methods.

\section{Acknowledgments}
A part of this research was supported by JSPS KAKENHI Grant Number 19K22863.

\bibliographystyle{aaai}
\bibliography{reference}

\end{document}